# Determination Of Structural Cracks Using Deep Learning Frameworks

Subhasis Dasgupta
Department of Data Science
Praxis Business School
Kolkata, INDIA
email: subhasisdasgupta1@acm.org

Jaydip Sen
Department of Data Science
Praxis Business School
Kolkata, INDIA
email: jaydip.sen@acm.org

Tuhina Halder
Dept. of Electronics and Communication Engg.
St. Thomas College of Engineering and Technology
Kolkata, INDIA
email: tuhina.halder@stcet.ac.in

*Abstract-* Structural crack detection is a critical task for public safety as it helps in preventing potential structural failures that could endanger lives. Manual detection by inexperienced personnel can be slow, inconsistent, and prone to human error, which may compromise the reliability of assessments. The current study addresses these challenges by introducing a novel deep-learning architecture designed to enhance the accuracy and efficiency of structural crack detection. In this research, various configurations of residual U-Net models were utilized. These models, due to their robustness in capturing fine details, were further integrated into an ensemble with a meta-model comprising convolutional blocks. This unique combination aimed to boost prediction efficiency beyond what individual models could achieve. The ensemble's performance was evaluated against well-established architectures such as SegNet and the traditional U-Net. Results demonstrated that the residual U-Net models outperformed their predecessors, particularly with low-resolution imagery, and the ensemble model exceeded the performance of individual models, proving it as the most effective. The assessment was based on the Intersection over Union (IoU) metric and DICE coefficient. The ensemble model achieved the highest scores, signifying superior accuracy. This advancement suggests way for more reliable automated systems in structural defects monitoring tasks.

I. INTRODUCTION

The structural integrity of buildings and infrastructure is essential to ensure public safety. A key aspect of this integrity is the quality of materials used during construction and the extent of curing, both of which contribute to the durability of structures. However, even well-constructed buildings and structures eventually develop cracks over time due to environmental factors, material fatigue, and general wear and tear. Such cracks can compromise stability, making the structures unsafe and potentially leading to failures if not addressed in time. In the past, many accidents have been linked to the failure to detect or assess cracks that have grown and propagated through structural elements. This oversight usually results in devastating consequences, signifying the importance of regular inspections and measures to detect cracks before they pose significant risks.

Detecting cracks in structures, however, is a challenging process. Traditionally, it has relied heavily on manual inspections, which are labor-intensive and subject to significant human error and variability. Not all inspectors possess the same skill level or experience, leading to inconsistencies in identifying critical faults through visual inspection. Moreover, the manual inspection process is often inefficient and costly, particularly for large or complex structures such as high-rise buildings, bridges, or industrial facilities. Inspecting tall structures poses additional risks and challenges, as working at height is both dangerous and uncomfortable for many workers. Similarly, inspecting structures in hazardous environments, such as chemical plants, presents life-threatening risks due to exposure to toxic substances. For these reasons, manual inspections are not only challenging but also pose considerable safety risks and financial costs.

To reduce these risks, drones are considered a valuable alternative in structural inspection. Operated remotely, drones can safely access difficult-to-reach areas and capture detailed visual data without posing life threats to humans. When equipped with cameras, drones can stream real-time video footage to experts on the ground, allowing them to visually assess structural health and identify cracks or other defects that require immediate intervention. This approach significantly reduces the risks associated with manual inspections, especially in hazardous environments or in the case of elevated structures. Moreover, recent advances in artificial intelligence and deep learning technologies present opportunities to further automate this process. Integrating a crack detection module with the drone's video feed, it becomes possible to analyze footage in real-time. This allows an AI-powered system to automatically assist human inspectors by identifying potential cracks. The combination of deep learning algorithms and high-performance graphic processing units (GPUs) enables rapid, real-time analysis of video data, thus enhancing the speed and accuracy of crack detection.

The current study explores various deep learning architectures specifically designed to improve crack detection accuracy. Crack detection can be approached as a task of object detection and masking, where specific model architectures are trained to identify cracks within video frames or images accurately. Additionally, this study proposes a novel architecture that uses an ensemble model to enhance performance further. By combining multiple models, each with distinct configurations and capabilities, the ensemble approach aims to combine the strengths of individual models. The process could yield a more robust and reliable crack detection system. The architecture was evaluated for its effectiveness in identifying cracks with high precision, offering a potential solution to improve structural safety and reduce the likelihood of undetected defects.

## II. RELATED WORKS

Researchers have investigated structural crack detection extensively, with methods generally falling into two main categories: object detection-based and semantic segmentation-based approaches. Object detection-based studies deal with models like Faster R-CNN, which has been employed in multiple works to identify cracks in structures [1], [2], [3], [4]. These studies utilize region proposal networks to highlight areas with potential cracks, allowing the models to focus on specific regions for better detection. Faster R-CNN has been noted for its ability to balance accuracy and efficiency. However, its effectiveness can be limited by complex textures and backgrounds in structural images. Another popular model used in object detection-based crack detection is Single-Shot Detection (SSD), which prioritizes real-time processing and operates well in environments where speed is critical. The SSD model has also been applied successfully in crack detection [5], [6], [7], though it may not reach the same level of precision as models that incorporate region proposals. Additionally, Mask R-CNN has gained attention in crack detection studies, as it enables pixel-level annotation by masking detected cracks, providing a finer level of detail in detection. This method has demonstrated significant improvements in predicting crack boundaries [8], [9], [10], [11], though it requires meticulous image annotation, which can be labor-intensive.

In contrast, semantic segmentation-based methods approach crack detection differently by focusing on pixel-level classification for the entire image. The UNet architecture is one of the most commonly used models in this approach. Designed initially for biomedical segmentation, UNet has a unique architecture that concatenates feature maps from the encoder with the corresponding layers in the decoder. This skip-connection mechanism enhances the model's capability to capture fine details, which is particularly beneficial when training on limited data, as is often the case in crack detection studies [12]. Following the success of the original UNet model, various modified versions of UNet have emerged in crack detection research [13], [14], [15], [16], [17]. These include architectures that incorporate residual blocks, which improve model learning by handling the issues related to vanishing gradients. Residual UNet architectures [18] had shown promising results in accurately detecting cracks with improved performance metrics in terms of precision and recall.

While these methods have produced valuable insights into structural crack detection, most of the studies focus on individual models without utilizing ensemble techniques. Ensemble methods, which combine predictions from multiple models to achieve a more robust output, have shown effectiveness in other domains but remain underexplored in crack detection research. In this study, the concept of ensemble learning is introduced by combining multiple UNet models with different configurations. A meta-model was used to integrate their predictions. This approach leverages the strengths of each individual model to create a more comprehensive and accurate crack detection system, which could enhance the overall predictive power.

## III. METHODOLOGY

The study follows a structured approach to analyze structural cracks using deep learning frameworks. The methodology is explained in the following subsections.

*Date Collection*: For any research data plays an important role. Several datasets related to structural defects (mainly cracks) are available on the internet. The images are available in both high and low resolution. Since the study dealt with deep learning, a dataset was needed that contains quite a large number of images with the associated masking of cracks. Hence, the CrackSeg9k [19] dataset was chosen as this dataset contains above 9k images of surfaces with cracks and no cracks. The dataset contains images all having resolutions 480 x 480 and the corresponding cracks were annotated with masks having the same resolution. The images were annotated using image processing techniques and hence the mask images were also not strictly binary in nature, meaning that the pixel values range from 0 to 1 without being strictly 0 or 1. The size of the data was approximately 1.4 GB and due to lower resolution, the dataset posed considerable challenges to train models to do segmentation of cracks in the images. Figure 1 shows a sample of the data with the associated masks. It can be seen clearly that the images are of low resolution. The masks were created using algorithms and hence, not every crack was identified correctly.

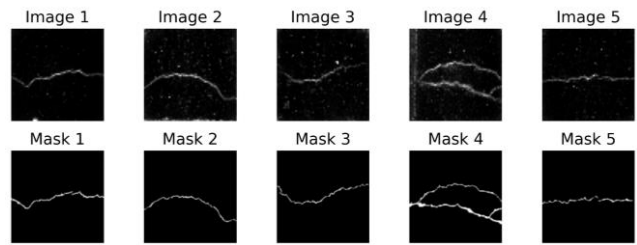

**Figure 1**: Sample images of the test dataset with the corresponding masks for prediction

*Data Preprocessing:* To reduce the memory footprint and time complexity, the raw images were converted to grayscale, which simplified the data structure by reducing each pixel from three color channels (RGB) to one. This resulted in fewer operations during the convolutional processing stage, allowing the convolution operations and weight updat-ing process to be executed more efficiently. This also ensured lesser time consumption and lesser utilization of GPU memory. The conversion to grayscale also helped in accelerating the overall training process since it limited the number of calculations required without a substantial loss in feature detection. Additionally, the images were resized to a resolution of 128 x 128 pixels to further optimize the computational workload. Reducing the image size reduced the number of pixels the neural network had to process, which directly lowered the demand on memory. These preprocessing steps were essential for handling large datasets efficiently, ensuring that the model could run smoothly on standard GPU hardware with limited resources. This approach made it feasible to train deep learning models even in constrained environments while maintaining model's performance. It also ensured that researchers and practitioners could work with larger datasets without facing prohibitive memory or processing constraints. Overall, these optimizations contributed to a more streamlined and accessible workflow for structural crack detection tasks.

*Model Architecture:* The dataset in this study was selected to specifically align with the task of image segmentation, and for this purpose, the Unet architecture proved to be highly effective. Unet's architecture incorporates skip connections to enhance the information flow during the decoding process, which helps in improving feature reconstruction. Later, SegNet was proposed for image segmentation, utilizing an autoencoder-like structure. The notable difference in SegNet is the incorporation of a VGG16-like architecture without fully connected layers and without skip connections, unlike Unet, which leveraged skip connections for feature mapping.

However, both Unet and SegNet lacked residual connections within their encoder and decoder components. This limitation was addressed by the development of the Residual Unet architecture, which integrated residual connections to improve model performance by allowing gradients to flow more effectively during backpropagation. In the present study, this concept was applied by constructing multiple shallow Residual Unet models, which were easier to train due to their simpler structure. Each of these shallow Residual Unets was trained independently on the training images. Once trained, their outputs were combined and passed through an additional sequence of convolutional layers to achieve a refined final outcome, enhancing the overall segmentation performance.

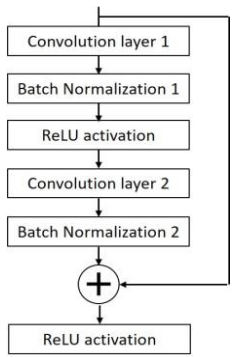

**Figure 2.** Residual connection used in residual Unet

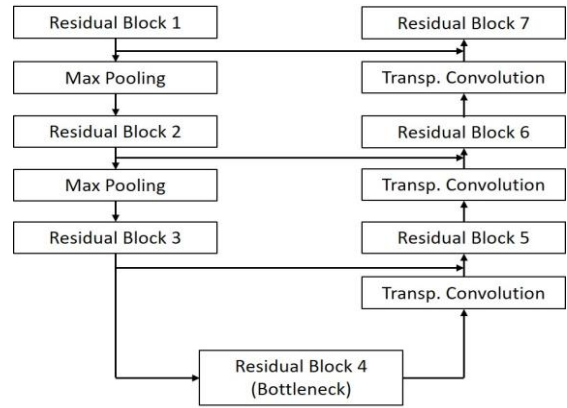

**Figure 3.** Residual Unet architecture with transpose convolution (shown as Transp. Convolution)

The residual connection is depicted in Figure 2, while the complete architecture of the Residual Unet is illustrated in Figure 3. The skip connections are denoted by horizontal arrows linking the outputs of the residual blocks within the encoder section to corresponding points in the decoder section following the transpose convolution operation. Transpose convolution was employed instead of traditional upsampling to enhance the processing and resolution of pixel values. Subsequently, leveraging the principles of the Residual Unet architecture, the proposed model was developed, as detailed in the block diagram presented in Figure 4.

The residual U-Nets used in this study featured an essential hyperparameter: kernel size. These kernels played crucial roles in feature identification and processing, influencing how each model interprets input images. To develop an effective ensemble structure, the kernel sizes were varied for different residual U-Net models. This variation enabled each model to process and analyze the input images in a unique manner, capturing distinct features and improving the ensemble's overall predictive capability. The combined outputs from these residual U-Net models were then passed through downstream convolutional blocks. These blocks functioned to further refine and fine-tune the mask prediction, enhancing the precision of the crack detection.

To ensure a thorough evaluation of the ensemble's performance, three key metrics were considered: validation loss, Intersection over Union (IoU), and the DICE coefficient. Validation loss provided a measure of how well the model generalized to unseen data during training, serving as an indicator of overfitting or underfitting. The IoU metric, commonly used in image segmentation tasks, measured the overlap between the predicted and actual crack regions, offering insight into the model's localization capabilities. The DICE coefficient, another critical metric, evaluated the accuracy of the predicted segmentation mask by considering both precision and recall. These comprehensive evaluations affirmed that the ensemble model outperformed standalone residual U-Nets and traditional architectures, demonstrating improved accuracy and reliability in structural crack detection.

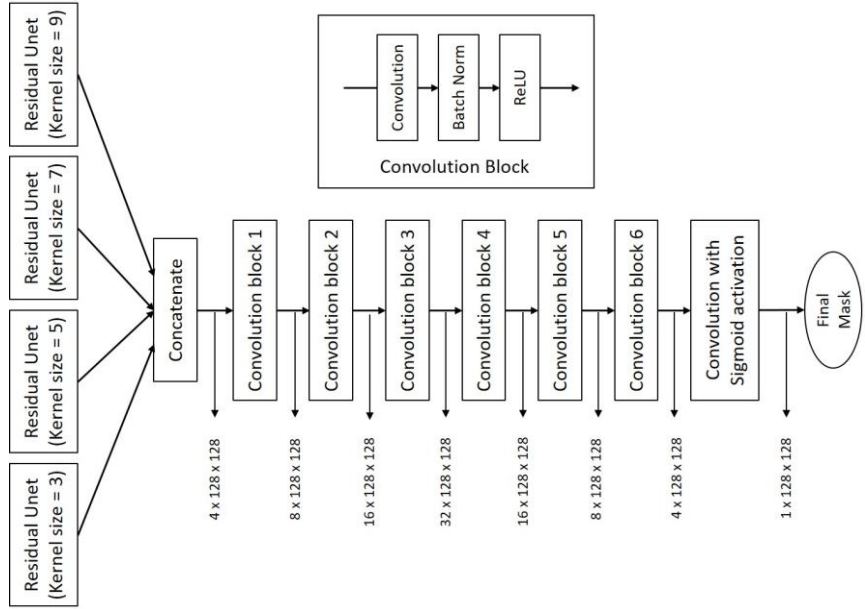

**Figure 4.** The proposed Ensemble Residual Unet architecture for image segmentation

## IV. RESULTS AND ANALYSIS

The training process involved two steps: first, training the Unets individually, and second, training the meta-model with a convolution block for fine-tuning. The residual Unets used in this study were shallow in depth, so initially, they were trained with the training samples to predict the masks individually. Later, the trained residual Unets were frozen so that, during the ensemble model training, only the convolution blocks were trained.

The models were trained using a batch size of 32 for 15 epochs using P100 GPU having 16 GB RAM and a system with 32 GB RAM. The training losses went down during the training process and after 10 epochs, almost all the model reached their respective saturation level. The loss function used was binary cross-entropy loss. The decrease in training loss is shown in Figure 5.

Since the images used in the training sample had low resolution, the Unet architecture without residual blocks found it very difficult to lower the training losses compared to other models. The SegNet model, due to its VGG16 like structure

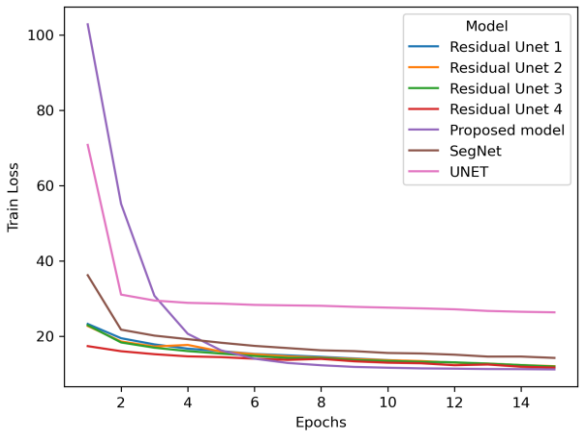

**Figure 5.** Movement of training loss of different models during the training process

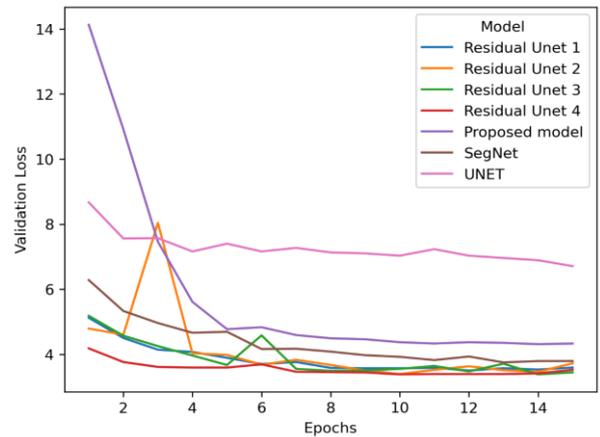

**Figure 6.** Movement of validation loss of different models during the validation process

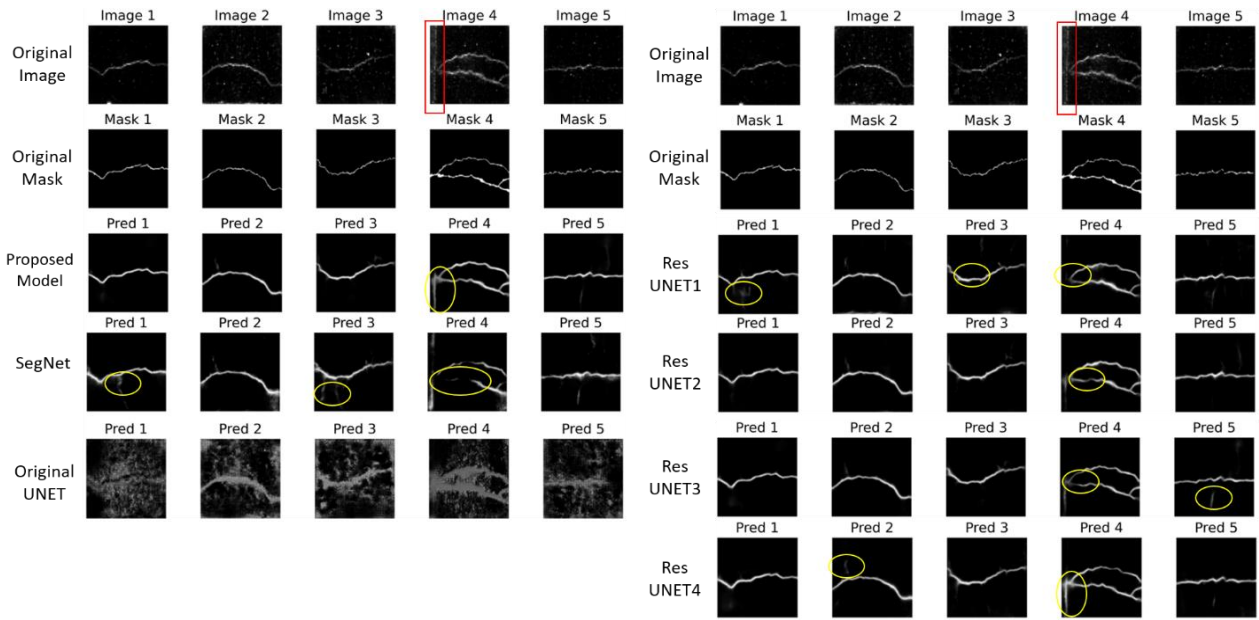

**Figure 7:** Performance of the 7 models on the sample dataset. The ovals show the major deviations from the original masks. The red rectangle shows that the original mask did not capture the crack properly

could identify the patterns much better than basic Unet architecture. The proposed model had higher loss in the beginning but it dropped down very quickly (as can be seen in Figure 5) before getting saturated. In fact, the training loss of the proposed model dropped down below the level of SegNet and Unet after 15 epochs. Lowering training loss only may also lead to overfit models and hence the models were evaluated with validation dataset also to understand whether they were getting overfit or not. Figure 6 shows the movement of validation losses of these models when the models were applied to the validation data.

Visual inspection of the models' performances is shown in Figure 7. The first two rows show the original images and the ground truth masks. The rows shown below these two rows are the prediction of the cracks by the models. It can be seen that the original UNET performed very poorly. SegNet performed better than UNET significantly but there were flaws in predictions as marked by the ovals. The residual UNETs were also giving better results but not without flaws. There predictions contained either breaks or false positive values. The proposed model performed much better in these respects. The extra tail which can be seen in the prediction of image 4 was mainly due to a missed crack as shown by the red box in the Figure 7.

It can be seen in Figure 6 that the proposed ensemble model has a higher validation loss than individual residual Unets and SegNet. This would usually mean that the ensemble model was not doing better than the individual model and was definitely inferior to the SegNet. However, it is to be understood that the models' performances are also to be judged based on established metrics such as Intersection over Union (IoU) and DICE coefficient rather than relying only on the reduction of losses. That is why the same models were applied to the test data, and this time three different measures were considered, i.e., test loss, IoU and DICE coefficient. The performances of the models are shown in Table 1 for comparative analysis. The test dataset had a lesser number of images than the training and validation sets, and hence, the losses also dropped compared to the validation process. Table 1 shows a clear picture that the ensemble model outperformed all the other models in terms of IoU and DICE coefficient. Performances of SegNet and Unet were far below the proposed network in terms of the IoU and DICE metric.

**Table 1.** Comparative analysis of the models on the test dataset

| Model | Test Loss | IoU | DICE Co-eff |
|---|---|---|---|
| Residual Unet 1 | **3.41** | 43.97% | 60.10% |
| Residual Unet 2 | 3.59 | 42.13% | 58.37% |
| Residual Unet 3 | 3.61 | 45.59% | 61.70% |
| Residual Unet 4 | 3.47 | 42.29% | 58.57% |
| **Proposed model** | 3.56 | **46.92%** | **63.00%** |
| SegNet | 3.59 | 39.86% | 56.08% |
| UNET | 4.99 | 10.68% | 17.65% |

## V. CONCLUSION

The present study tried to detect structural cracks using deep learning architecture. In this study, residual Unets were trained with different kernels size to analyze the images differently from each other and a new ensemble model with convolution blocks as meta model was used to combine the predictions of all the residual Unets in an end-to-end manner. For comparison purposes, SegNet and basic Unet architectures were also trained on the same training set. It was seen that the residual Unets were performing much better than the SegNet and Unet when applied to low-resolution images. Not only that, the proposed ensemble residual Unet performed better than individual residual Unets on the test data. This suggests that the ensemble model can learn different aspects from different predictions and intelligently combine the outcomes to give a better prediction than the individual models. In future, a similar model can be applied in other domain areas to judge the efficacy of the model in different domains.